\documentclass[letterpaper, 10 pt, conference]{ieeeconf}  % Comment this line out if you need a4paper

\IEEEoverridecommandlockouts                              % This command is only needed if
\usepackage{amsmath,amssymb,amsfonts}
\usepackage{algorithmic}
\usepackage{array}
\usepackage{textcomp}
\usepackage{stfloats}
\usepackage{url}
\usepackage{verbatim}
\usepackage{graphicx}

\usepackage{cite}
\usepackage{caption}
\usepackage{tikz}
\usepackage{etoolbox}
\usepackage{booktabs}
\usepackage{color}

\usepackage{multirow}

\usepackage[mathscr]{eucal}

\title{\LARGE \bf
RCGNet: RGB-based Category-Level 6D Object Pose Estimation with  Geometric Guidance
}

\author{Sheng~Yu,
        Di-Hua~Zhai$^\ast$,
        and Yuanqing~Xia,~\IEEEmembership{Fellow,~IEEE}
\thanks{The work was supported by the National Natural Science Foundation
of China under Grant 62173035, Grant 61803033 and Grant 61836001. \textit{(Corresponding author: Di-Hua Zhai.)}}
\thanks{Sheng Yu and Di-Hua Zhai are with School of Automation, Beijing Institute of Technology,
Beijing 100081, China, Email: yusheng@bit.edu.cn, zhaidih@bit.edu.cn.}% <-this % stops a space
\thanks{Yuanqing Xia is with Zhongyuan University of Technology, Zhengzhou
450007, Henan, China, and also with School of Automation, Beijing Institute of Technology,
Beijing 100081, China, Email: xia$\_$yuanqing@bit.edu.cn.}
}

\begin{document}

\maketitle
\thispagestyle{empty}
\pagestyle{empty}

%%%%%%%%%%%%%%%%%%%%%%%%%%%%%%%%%%%%%%%%%%%%%%%%%%%%%%%%%%%%%%%%%%%%%%%%%%%%%%%%
\begin{abstract}
While most current RGB-D-based category-level object pose estimation methods achieve strong performance, they face significant challenges in scenes lacking depth information. In this paper, we propose a novel category-level object pose estimation approach that relies solely on RGB images. This method enables accurate pose estimation in real-world scenarios without the need for depth data. Specifically, we design a transformer-based neural network for category-level object pose estimation, where the transformer is employed to predict and fuse the geometric features of the target object. To ensure that these predicted geometric features faithfully capture the object's geometry, we introduce a geometric feature-guided algorithm, which enhances the network's ability to effectively represent the object's geometric information. Finally, we utilize the RANSAC-PnP algorithm to compute the object's pose, addressing the challenges associated with variable object scales in pose estimation. Experimental results on benchmark datasets demonstrate that our approach is not only highly efficient but also achieves superior accuracy compared to previous RGB-based methods. These promising results offer a new perspective for advancing category-level object pose estimation using RGB images.
\end{abstract}

\section{Introduction}
6D object pose estimation has been widely used in many areas such as robotic grasping \cite{wang2019densefusion}, autonomous driving \cite{8053815} and augmented reality \cite{su2019deep}. Instance-level object pose estimation has been firstly widely investigated, such as \cite{zhou2023deep,lin2024hipose,hong2024rdpn6d}. They determine the scale information and texture information of the object based on the 3D model of the object as well as the RGB-D image, which in turn performs 6D pose estimation. Although instance-level object pose estimation methods achieve good results in datasets and some complex scenes, these methods heavily rely on the 3D model of the object. When faced with unknown objects in unknown scenes, these methods struggle to complete the object pose estimation task.

To address this issue, researchers propose a variety of category-level object 6D pose estimation methods \cite{wang2019normalized,tian2020shape,chen2021sgpa}. Unlike instance-level pose estimation methods, category-level object pose estimation doesn't require a 3D model of the object and can predict the pose of unknown objects within known categories. However, typically depend on depth information, which limits their applicability in scenarios where such data is unavailable.

Therefore, researchers propose some RGB-based category-level object pose estimation methods, such as \cite{lee2021category,fan2022object,wei2024rgb,zhang2024lapose}. These methods can perform a certain degree of object pose estimation when depth information is unavailable. However, estimating poses solely based on RGB images is highly challenging due to the lack of depth information. To solve the problem of missing depth information, some methods, such as  \cite{lee2021category,fan2022object,wei2024rgb}, require predicting a depth image each time pose estimation is performed before completing the object pose estimation, which makes the process more cumbersome. Although other methods, such as \cite{zhang2024lapose}, suggest different approaches, the network still requires additional training of a scale network to achieve accurate pose estimation. The network training process is relatively cumbersome.

To address the issues with previous methods, in this paper, we attempt to build a category-level object pose estimation network with a simple training process that can directly predict the geometric features of objects. In recent years, vision transformers have achieved excellent results. Compared to CNNs, transformers have stronger learning capabilities. In this paper, we use DINOv2 \cite{oquab2023dinov2} as the backbone to extract features from RGB images and predict the geometric features of objects. The advantage of this approach is that the network does not need to predict depth images first before predicting geometric features, and it can directly rely on RGB images to complete the prediction of object geometric features, thus eliminating the cumbersome process.

However, considering the significant difference between the information contained in RGB images and depth information, directly using RGB images for prediction without imposing constraints may lead to large deviations between the predicted geometric features and the actual geometric features, resulting in a decline in pose estimation accuracy. This issue has not been considered in previous related methods. If we can impose certain constraints during the training process to guide the RGB-based predicted features toward the actual features, it can effectively improve the accuracy of pose estimation. 

To address this issue, we propose a geometric feature guidance method in this paper, where real geometric features are used as labels to guide the predicted geometric features to approach the real ones, until the real geometric features can be accurately predicted. In this way, we can not only eliminate the additional computational overhead of predicting depth images but also obtain the geometric features of objects in a relatively simple and effective manner. More importantly, during the pose prediction process, we can reduce the network's dependence on depth information to some extent, achieving pose estimation using only RGB images, which provides a new approach for future methods.

Since we cannot directly obtain the depth information of objects, making it difficult to determine the scale of objects, we need to decouple the object pose from the object scale. During the pose estimation process, we additionally construct a scale estimation branch to estimate the size of the object. Then, we use the RANSAC-PnP algorithm to solve for the object pose, obtaining the final pose of the object. In the training process, we achieve this in one step without the need for additional adjustments to the object scale during training, unlike methods \cite{zhang2024lapose} that require such adjustments, which would make the training process cumbersome. Our entire network is highly efficient and achieves performance surpassing previous methods on the CAMERA25 and REAL275 datasets \cite{wang2019normalized}.

In summary, the main contributions of this paper are:
\begin{enumerate}
\item We propose a new RGB-based category-level object pose estimation network with vision transformer called RCGNet, which does not require additional predictions of depth images or other related images during the pose estimation process.

\item We propose a geometric guidance method to guide the network in efficiently predicting the geometric

\item We train and test RCGNet on the CAMERA25 and REAL275 datasets. Experimental results show that, compared to previous methods, RCGNet can effectively complete object pose estimation and achieve better pose estimation results.

\end{enumerate}

\section{Related Works}

\subsection{Instance-Level Object Pose Estimation}
Recently, with the development of deep learning, a series of deep leaning-based instance-level object pose estimation methods have been proposed, such as \cite{xiang2018posecnn,wang2019densefusion,peng2019pvnet,tekin2018real}. These methods perform object pose estimation using RGB images or RGB-D images. For RGB-based methods, they often complete object pose estimation through direct regression \cite{xiang2018posecnn} or 2D-3D keypoint correspondence \cite{peng2019pvnet}. For example, in \cite{xiang2018posecnn}, Xiang et al. propose PoseCNN, which takes RGB images as input, and directly outputs the 6D pose of objects based on the features extracted by VGG-19. In \cite{peng2019pvnet}, PVNet detects keypoints in the RGB images and uses PnP to solve the object pose. For RGB-D image-based methods, they can also rely on direct regression \cite{wang2019densefusion} and keypoint correspondence \cite{he2020pvn3d} to complete object pose estimation. However, compared to RGB image-based methods, RGB-D image-based methods can fully utilize the geometric information provided by the depth image, enabling more accurate object pose estimation. Although instance-level object pose estimation achieves good results, these methods rely on the 3D model of the object. When facing unknown objects, such methods are difficult to use in practice.

\subsection{Category-Level Object Pose Estimation}
%Although instance-level object pose estimation methods have got much progress, they heavily rely on the 3D model of the objects, and are hard to be applied to other novel objects. 
To solve the problem in instance-level object pose estimation, researchers have proposed category-level object pose estimation methods, such as \cite{wang2019normalized,
chen2021sgpa,lin2021dualposenet,tian2020shape,lin2023vi,chen2024secondpose,lin2024instance}. In \cite{wang2019normalized}, Wang et al. first propose Normalized Object Coordinate Space (NOCS). All objects are normalized in a same coordinate space, and objects belonging to the same category have the same coordinate orientation. The 6D pose is calculated by predicting the NOCS model of the object.  In \cite{tian2020shape}, Tian et al. propose a novel pose estimation method based on prior point cloud. The prior point cloud provides more information of the objects in the same category, and help improve the accuracy of pose estimation. In \cite{lin2024instance}, Lin et al. propose a new object pose estimation method with instance-adaptive keypoints.

Although these methods achieve good results in category-level object pose estimation, they all rely on the object's depth information. In certain extreme environments where depth information is unavailable, these methods cannot complete the object pose estimation task. To address this, some researchers attempt to perform object pose estimation using only RGB images, such as \cite{chen2020category,lee2021category,fan2022object,wei2024rgb,zhang2024lapose}. In \cite{fan2022object}, Fan et al. predict object pose via Umeyama algorithm from the 3D-3D correspondences established by NOCS coordinates prediction. In \cite{wei2024rgb}, Wei et al. estimate object pose with predicted object normals and relative depths, and solve the object pose with 2D-3D correspondences. In \cite{zhang2024lapose}, Zhang et al. propose a novel object pose estimation method with Laplacian mixture model. Although these methods can complete object pose estimation, the accuracy of pose estimation still needs further improvement.

\begin{figure*}[!h]
	\centering
	\includegraphics[width=0.9\linewidth]{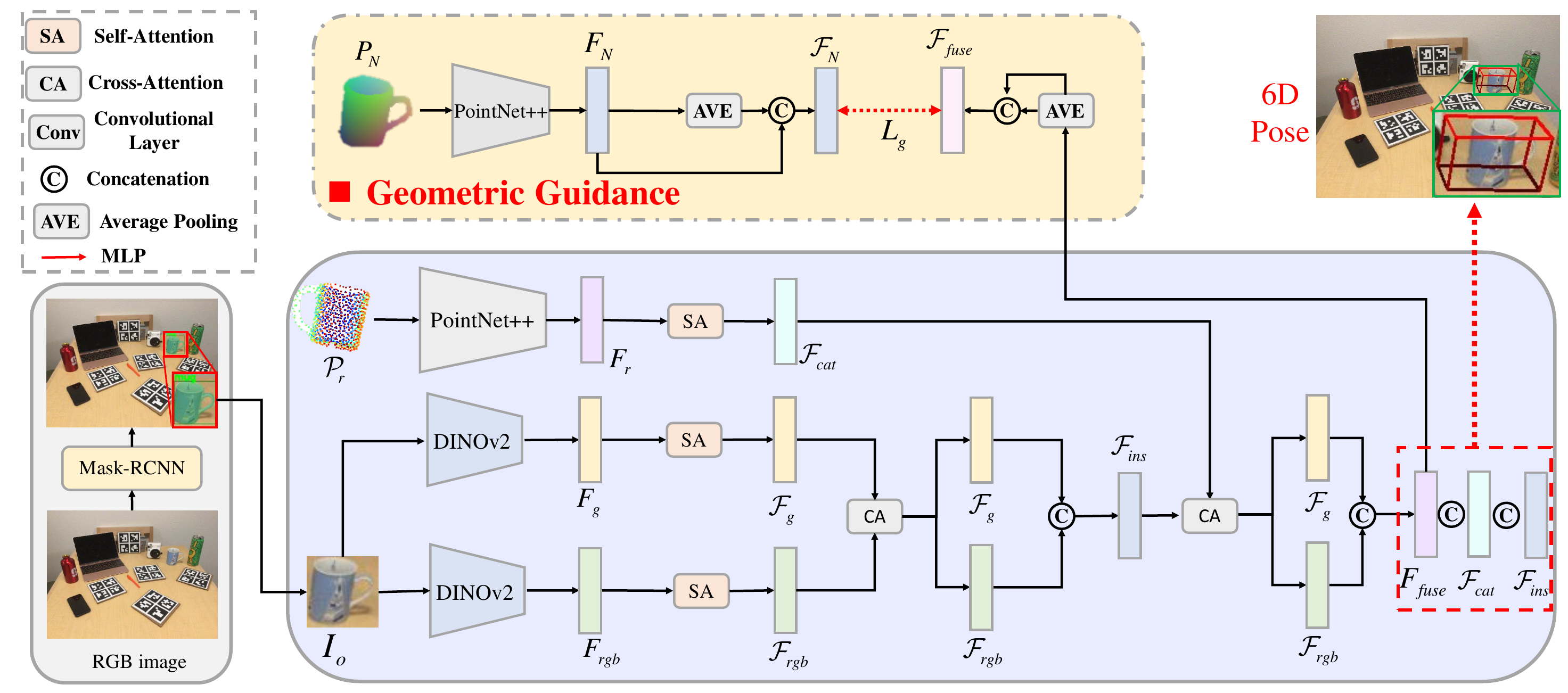}
	\caption{The pipeline of the RCGNet.}
	\label{fig1}
\vspace{-0.5cm}
\end{figure*}

\section{Approach}

\subsection{Pipeline Overview}

The pipeline of the training process of RCGNet is shown in Fig.\ref{fig1}. We take the RGB image and the prior point cloud as the input, and use a instance segmentation network, such as Mask-RCNN \cite{he2017mask}, to segment the object mask from the image. We crop the RGB image using the instance mask to obtain RGB patches of the target object. 

We define $\mathcal{P}_{r}\in \mathbb{R}^{N_{r}\times 3}$ as the prior point cloud of the target object, which contains $N_{r}$ points, and $I_{o}\in  \mathbb{R}^{H\times W\times 3}$ is the RGB image of the object. We use DINOv2 \cite{oquab2023dinov2} as the backbone to extract the features in the RGB images. DINOv2 can extract semantic-aware information from RGB images that can be well leveraged to establish zero-shot semantic correspondences, rendering it an excellent method for rich semantic information extraction. We also utilize the PointNet++ \cite{qi2017pointnet++} as the backbone to extract the features in the prior point cloud.

\subsection{Geometric Embedding and Feature Fusion}

First, we predict the geometric features of the object using RGB images. In this paper, unlike other methods, we directly use the object's RGB image to predict its geometric features. To ensure that the geometric features of the object are learned effectively, we use a transformer-based DINOv2 as the backbone of the network. Compared to traditional convolutional networks, transformers can more effectively extract and learn features from images, and they can enhance the network's attention to key features, achieving better prediction results.

With the help of DINOv2, we perform feature extraction from RGB images and predict the geometric features. Based on the characteristics of DINOv2, we freeze the weights of DINOv2 throughout the training process. As a result, we obtain the RGB image features $F_{rgb}$ and the object's geometric features $F_g$.

We utilize MLP layers to generate the three parameters of the self-attention transformer module,
\begin{equation}
q_{*}^{S},k_{*}^{S},v_{*}^{S}=\text{MLP}(F_{*})
\end{equation}
where $* \in \{rgb,g\}$, $q,k,v$ indicate \textit{query, key,} and \textit{value} respectively, $S$ indicates the parameter used in the self-attention module.

Then we utilize the self-attention module to perform feature extraction and attention adjustment
\begin{equation}
\mathcal{F}_{rgb}=\text{SA}(q_{rgb}^{S},k_{rgb}^{S},v_{rgb}^{S})+F_{rgb},\mathcal{F}_{g}=\text{SA}(q_{g}^{S},k_{g}^{S},v_{g}^{S})+F_{g}
\end{equation}
where SA indicates a self-attention module, which can be calculated by
\begin{equation}
\text{SA}(q,k,v)=softmax\left(\frac{q\times k^{T}}{\sqrt{d_{k}}}\right)v
\end{equation}
where $d_{k}$ is the dimension of $k$.

To aggregate RGB and geometric features, we utilize transformer-based cross-attention in this paper. Firstly, we generate \textit{query, key,} and \textit{value} of the cross-attention module,
\begin{align}
q_{*}^{C},k_{*}^{C},v_{*}^{C}&=\text{MLP}(\mathcal{F}_{*})
\end{align}
where $* \in \{rgb,g\}$, $C$ indicates the parameter will be used in the cross-attention. Then we calculate the cross-attention of the RGB features and geometric features, and perform feature fusion
\begin{align}
\mathcal{F}_{rgb}&=\text{CA}_{rgb}(q_{rgb}^{C}, k_{g}^{C},v_{g}^{C}), \\
\mathcal{F}_{g}&=\text{CA}_{g}(q_{g}^{C},k_{rgb}^{C},v_{rgb}^{C})
\end{align}
where $\text{CA}_{*}$ indicates the cross-attention. The calculation process is indicated as
\begin{equation}
\text{CA}=softmax\left(\frac{q_{i}^{C}\times (k_{j}^{C})^{T}}{\sqrt{d_{k}}}\right)v_{j}^{C}+q_{i}^{C}
\end{equation}
where $i,j \in \{rgb,g\}$ and $i\neq j$.

We concatenate $\mathcal{F}_{rgb}$ and $\mathcal{F}_{g}$, and get the instance fusion feature $\mathcal{F}_{ins}$.

Due to the lack of point clouds for the object, in RCGNet, we also use the prior point cloud of the object's corresponding category as input and employ PointNet++ as the backbone to perform feature extraction from the prior point cloud, obtaining $F_r$. We also use the SA module to extract the category feature in $F_r$, and get $\mathcal{F}_{cat}$.

To effectively utilize both the instance features and category features of the object, we also use CA to fuse them, obtaining the fused feature $\mathcal{F}_{fuse}$.

\subsection{Geometric Guidance}

Although we effectively complete the network's prediction and fusion of object geometric features, this feature prediction is one-way. The network does not impose any labels during the training process to restrict and guide the prediction of these geometric features. This leads to poor prediction results for geometric features, which affects the accuracy of pose estimation. This key point has been overlooked in previous methods.

To solve this problem, in this paper, we propose a geometric guidance method to help the network learn the prediction of object geometric features during the training process. It should be noted that this prediction process is only used during training and does not involve any point cloud information during the prediction process.

Since we predict the object's pose using the NOCS coordinate information, we transfer the object's point cloud to NOCS to obtain the object's point cloud in NOCS
\begin{equation}
P_{N}=(P_o-t)/s \times R
\end{equation}
where $P_{N}\in \mathbb{R}^{N_{o}\times 3}$ is the point cloud in NOCS, $P_o\in \mathbb{R}^{N_{o}\times 3}$ is the object point cloud, $R,t,s$ indicate the ground truth rotation, translation and size of the object respectively.

Then, we use PointNet++ to extract features from $P_{N}$, obtaining the object's NOCS geometric features $F_N$. Since relying directly on $F_N$ can effectively complete the object's pose estimation, we hope the fused features of the object can effectively approach the real NOCS geometric features.
We perform average pooling on $F_{fuse}$ to reduce it to 1D, then repeat the dimension and concatenate it with $F_{fuse}$
\begin{equation}
\mathcal{F}_{fuse}=\copyright(F_{fuse},\text{M}(F_{fuse}))
\end{equation}
where $\copyright(,)$ indicates feature concatenation, M indiactes average pooling.

After that, we construct an MSE loss function to make $\mathcal{F}_{fuse}$ and $F_N$ as close as possible, achieving geometric guidance $L_{g}=\text{MSE}(\mathcal{F}_{fuse},F_N)$.

Based on the subsequent experimental results, we can see that this operation method is highly efficient and effectively guides the fused features to predict the geometric characteristics of the object. 

\subsection{Pose Estimation}
We rely on the fused features, using both local and global features, to predict the prior point cloud deformation offsets $D$, transformation matrices $A$
\begin{align}
A&=\text{MLP}(\copyright(\mathcal{F}_{ins},F_{fuse})) \\
D&=\text{MLP}(\copyright(\mathcal{F}_{cat},F_{fuse})) 
\end{align}
where MLP indicates MLP layers.

Due to the lack of depth information of objects, we find it difficult to directly predict the scale information of objects. To address this, we construct a benchmark scale $s_b$ for each category to prevent large fluctuations during the scale prediction process, and predict the scale variation of the object $\Delta S$. The predicted scale $s$ can be calculated by 
\begin{equation}
\Delta S=\text{MLP}(\copyright(\mathcal{F}_{ins},F_{fuse})) , s=s_b+s_b\times \Delta S
\end{equation}

Finally, we use the RANSAC-based PnP to calculate the object pose 
\begin{equation}
R,t=\text{PnP}(P_I,s\times A(P_r+D))
\end{equation}
where $P_I$ indicates the corresponding 2D pixel coordinates on image patch, $R,t$ indicate the rotation and translation.

\subsection{Loss Function}

We first calculate the scale loss function of the object
\begin{equation}
L_{s}=||\Delta \hat{S}-\Delta S||, \Delta \hat{S}=\frac{\hat{s}-s_b}{s_b}
\end{equation}
where $\hat{s}$ indicates the ground truth scale.

Then we calculate the NOCS correspondences loss $L_{corr}$ following \cite{chen2021sgpa}, which is used to supervise the prediction of $A$ and $D$.

Finally, we combine $L_g$ and calculate the network's final loss function
$L=\lambda_1 L_s+\lambda_2 L_{corr}+\lambda_3 L_g$.

\section{Experiments}

\subsection{Dataset}
For the category-level object pose estimation, the REAL275 dataset and CAMERA25 dataset are the benchmark datasets, which are proposed in \cite{wang2019normalized}. The CAMERA25 dataset contains 300K images, where 25K images are used for evaluation. The CAMERA25 dataset is generated by rendering the synthetic objects into real scenes. The REAL275 dataset is captured by a camera in the real world. The REAL275 dataset contains 4300 real-world images of 7 scenes used for training, and contains 2750 real-world images of 6 scenes for evaluation. Both two datasets mainly contain 6 different categories of objects: bottle, bowl, camera, can, laptop and mug.

\subsection{Training Details}
We perform the experiments on an NVIDIA GeForce RTX 3090 GPU, and the operating system is Ubuntu 18.04. We use PyTorch 1.12.1 as the deep learning framework, and CUDA 11.3 is used to accelerate the training process. We set the batch size as 16, and train the network for 120 epochs.

We process the dataset following the procedures in \cite{tian2020shape}, and use a instance segmentation network, \textit{e.g.,} Mask-RCNN, to perform the object detection and segmentation. %Based on the RGB-D images and intrinsic matrix of the camera, we generate the point cloud of the scene, and randomly select $N_o=1024$ points of each object in the point cloud.
We set $\lambda_1=1.0, \lambda_2=0.1$ and $\lambda_3=100.0$. The size of the cropped RGB image in the training process is $224\times 224$, and the number of the point in the prior point cloud is $N_r=1024$.

\subsection{Evaluation Metrics}
We follow the widely used evaluation metrics in \cite{fan2022object,wei2024rgb} to evaluate the performance of RCGNet. We use the 3D Intersection-Over-Union (IoU) with thresholds of 0.5, and 0.75 to jointly evaluate rotation and translation. To directly evaluate the rotation and translation, we also use $10cm$, $10^{\circ}$, and $10^{\circ}10cm$ to evaluate the performance of the network. If the errors blow the thresholds, we can judge this prediction is correct.
Based on these evaluation metrics, we will use overall mAP to evaluate the performance of RCGNet with other related SOTA methods.

\begin{figure*}[!h]
	\centering
	\includegraphics[width=1\linewidth]{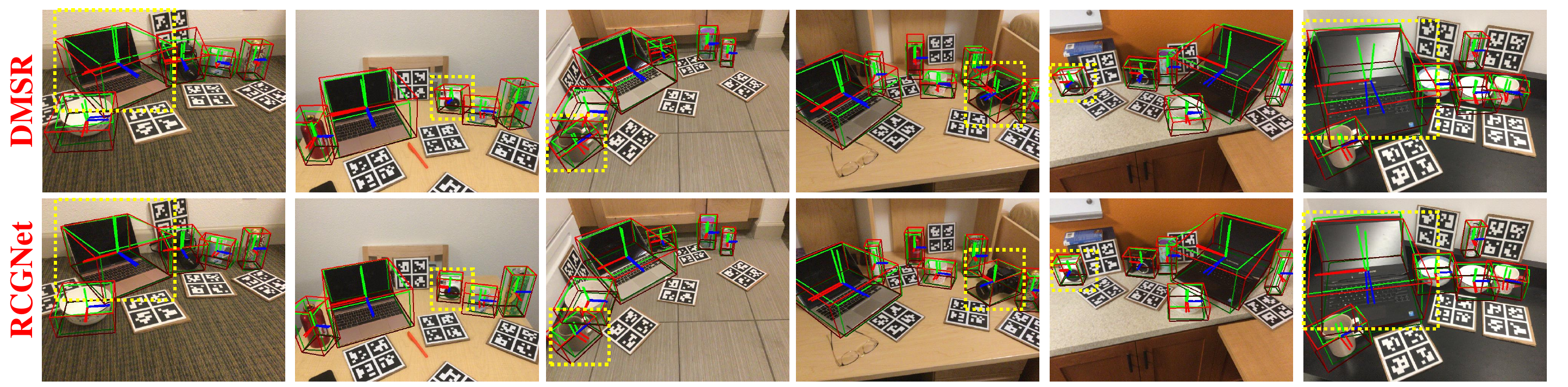}
	\caption{Qualitative results of DMSR and RCGNet. The bounding box in green line is the ground truth, and the red line is the prediction. The yellow dashed box indicates objects with pose differences.
}
	\label{fig2}
\vspace{-0.5cm}
\end{figure*}

\subsection{Comparison with State-of-the-Art Methods}
We compare the performance of RCGNet with recent state-of-the-art RGB-based methods on CAMERA25 dataset and REAL275 dataset, results are provided in TABLE \ref{compare}. We evaluate these methods using the $IoU_{50}$, $IoU_{75}$, $10cm$, $10^{\circ}$ and $10^{\circ}10cm$ metric.

\begin{table}[!h]
  \centering
\scriptsize
 \caption{Comparison with state-of-the-art methods on CAMERA25 dataset and REAL275 dataset. } \label{compare}
\setlength{\tabcolsep}{0.1mm}
  \begin{tabular}{c|c|c|c|c|c|c|c|c|c|c}
\hline
     \multirow{2}{*}{Method}   &\multicolumn{5}{c|}{CAMERA25} &\multicolumn{5}{c}{REAL275} \\
  \cline{2-11}
  &$IoU_{50}$ &$IoU_{75}$ &$10cm$  &$10^{\circ}$ &$10^{\circ}10cm$  &$IoU_{50}$ &$IoU_{75}$ &$10cm$  &$10^{\circ}$ &$10^{\circ}10cm$  \\
\hline

	Synthesis\cite{chen2020category}  &- &- &- &- &- &- &- &34.0 &14.2 &4.8 \\
MSOS\cite{lee2021category} &32.4 &5.1 &29.7 &60.8 &19.2 &23.4 &3.0 &39.5 &29.2 &9.6 \\
OLD-Net\cite{fan2022object} &32.1 &5.4 &30.1 &74.0 &23.4 &25.4 &1.9 &38.9 &37.0 &9.8 \\
DMSR\cite{wei2024rgb} &34.6 &6.5 &32.3 &81.4 &27.4 &28.3 &6.1 &37.3 &59.5 &23.6 \\
\hline
Ours   &\textbf{41.8} &\textbf{8.2}  &\textbf{39.5} &\textbf{87.2} &\textbf{32.8}  &\textbf{35.4} &\textbf{8.1} &\textbf{46.1} &\textbf{64.3} &\textbf{30.1}
\\
\hline
  \end{tabular}
\vspace{-0.5cm}
\end{table}

From the experimental results, we can see that, compared to previous methods, RCGNet significantly outperforms them on the dataset, achieving excellent prediction results and improving performance across multiple evaluation metrics. Previous methods, such as OLD-Net and DMSR, require additional computational resources for depth information prediction and rely on depth information to complete pose estimation. This leads to inaccurate pose estimation when there are deviations in the depth information, affecting the accuracy of pose estimation. In contrast to previous methods, we do not rely on additional depth information or surface normal vector information; we can directly predict the object's geometric information using only RGB images, which is much simpler and more straightforward. Additionally, we have supervision during the training process of geometric feature prediction, effectively guiding the prediction process and reducing the likelihood of large-scale deviations.

To further demonstrate the performance of pose estimation, we also provide the visualization results on the REAL275 dataset, as shown in Fig.\ref{fig2}. In the experiment, we compare DMSR and RCGNet, selecting multiple real-world objects in actual scenes for the pose estimation task. According to the pose estimation results, compared to DMSR, the method we propose can more effectively estimate the pose of complex objects, such as cameras, and achieve better pose estimation results. This is mainly due to the geometric guidance algorithm we designed, which, through effective geometric guidance, helps the network learn the geometric features of the target object and improves the accuracy of pose estimation.

\subsection{Ablation Studies}
To verify the effectiveness of the algorithm we propose, we conduct a series of ablation experiments on the REAL275 dataset. The experimental results are shown in TABLE \ref{abla}. 
\begin{table}[!h]
  \centering
\scriptsize
  \caption{Ablation studies  on REAL275 Dataset.} \label{abla}
\setlength{\tabcolsep}{1mm}
  \begin{tabular}{c|c|c|c|c|c|c|c}
\hline
  \multicolumn{3}{c|}{Modules} &\multicolumn{5}{c}{mAP(\%)} \\
\hline
Feature Fusion  &Guidance &Scale  &$IoU_{50}$ &$IoU_{75}$  &$10cm$ &$10^{\circ}$ &$10^{\circ}10cm$  \\
\hline
 \checkmark	& \checkmark	& &22.6	&3.4		&31.4		&51.8	  &18.9  \\	

 	\checkmark &	&\checkmark	&26.9		&5.9		&38.5		&57.6	 &22.4  \\		
 	&\checkmark		&\checkmark &29.7	&6.8		&41.2		&60.3  &25.6  \\		
\hline	
\checkmark 	&\checkmark	&\checkmark	&\textbf{35.4} &\textbf{8.1} &\textbf{46.1} &\textbf{64.3} &\textbf{30.1}   \\		
\hline
  \end{tabular}
\vspace{-0.3cm}
\end{table}

%First, when we add the transformer feature fusion, geometric guidance, and scale prediction to the network, the network achieves the best performance. Then, when we remove the scale module and predict directly through regression without relying on the benchmark scale, the experimental results are shown in the first row of the table. It can be observed that after removing the scale module, the network's prediction performance decreases, leading to less accurate scale prediction and significant variations in object scale prediction, which affects the accuracy of pose estimation.
First, incorporating transformer feature fusion, geometric guidance, and scale prediction into the network yields the best performance. Removing the scale module and directly predicting via regression without relying on the benchmark scale, as shown in the first row of the table, results in decreased prediction performance. This leads to less accurate scale predictions and significant variations in object scale, which negatively impacts pose estimation accuracy.

%Next, we explore the effectiveness of the geometric guidance module, and the experimental results are shown in the second row of the table. When the geometric guidance module is removed, the network can only rely on the existing knowledge to predict the object's geometric features. Due to the lack of effective geometric guidance, the network experiences biases in predicting geometric features, which impacts the pose prediction performance. At the same time, we can also observe that after removing the geometric guidance module, the network's pose estimation performance decreases significantly, highlighting the importance and effectiveness of geometric guidance in pose estimation.
Next, we examine the effectiveness of the geometric guidance module, with results shown in the second row of the table. Without the geometric guidance, the network relies solely on existing knowledge for predicting object features. This lack of effective guidance introduces biases in the predictions, impacting pose performance. The significant drop in pose estimation accuracy further highlights the importance and effectiveness of geometric guidance.

Finally, we test the effectiveness of the transformer feature fusion module. In the experiment, we replace the CA module with direct feature concatenation. According to the experimental results, it can be seen that when the CA module is removed, the network's performance decreases to some extent. This is mainly because the transformer plays an important role in adjusting feature attention and promoting feature complementarity. After its removal, the network lacks effective feature processing capabilities, leading to a decline in pose estimation accuracy.

\subsection{Experiments on Wild6D Dataset}

\begin{figure}[!h]
\vspace{0.15cm}
	\centering
	\includegraphics[width=1\linewidth]{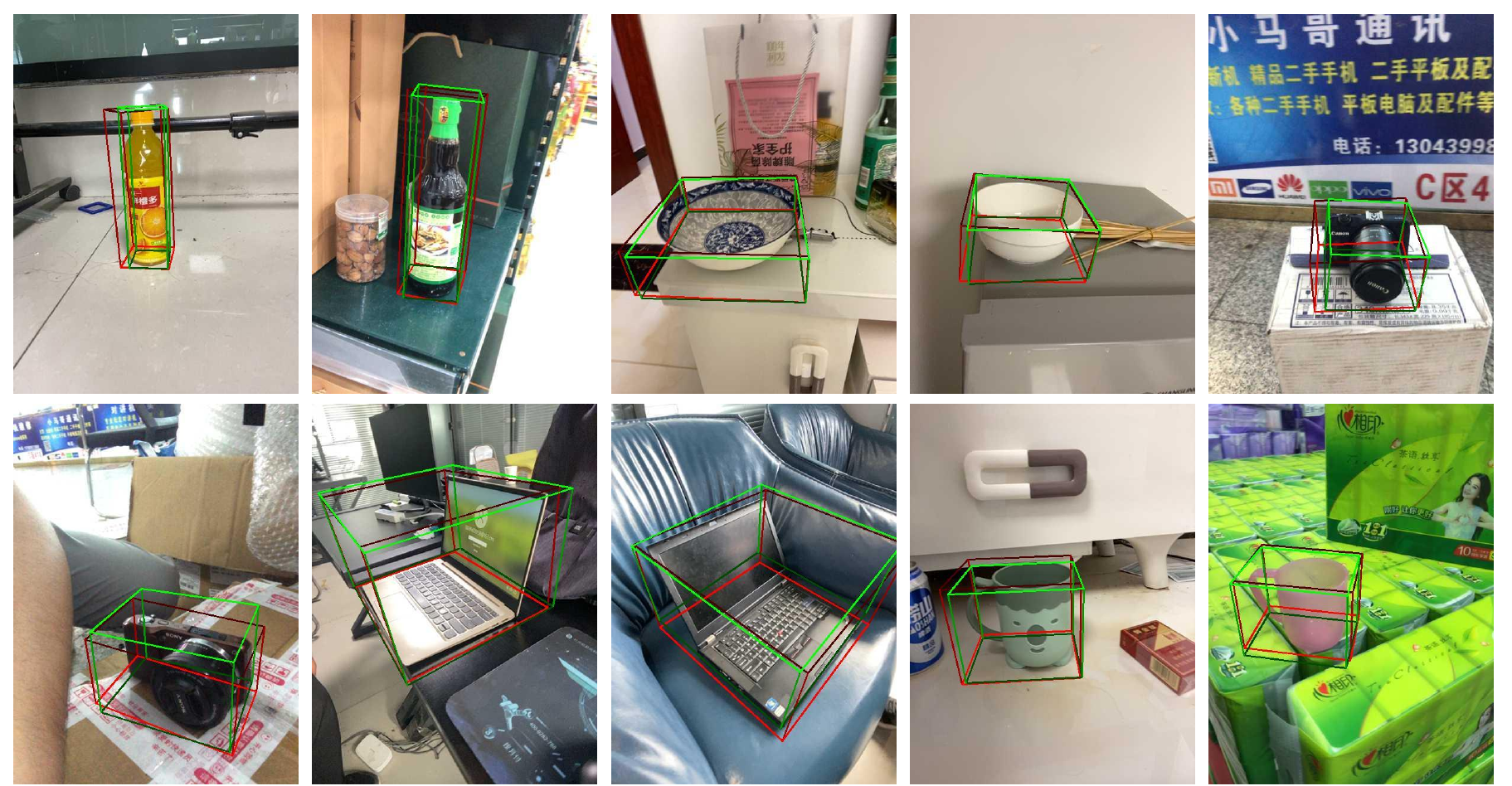}
	\caption{Qualitative results on Wild6D dataset. The bounding box in green line is the ground truth, and the red line is the prediction. 
}
	\label{fig8}
\vspace{-0.6cm}
\end{figure}

To validate the generalization capability of our designed network, we conduct pose estimation on the Wild6D dataset \cite{ze2022category}. In the Wild6D dataset, all objects are unknown and randomly placed in various real-world environments to test the network's generalization ability.
 We train RCGNet on the REAL275 and CAMERA25 datasets and perform testing on the Wild6D dataset, and qualitative visualization results are provided in Fig.\ref{fig8}.  The experimental results show that RCGNet can effectively generalize to pose estimation tasks for various unknown objects in real-world scenarios, achieving good results with high generalization and reliability.

\subsection{Discussions}
Although our method can effectively perform object pose estimation, it also has some limitations. First, our pose estimation relies on accurate object segmentation. When there are deviations in the predicted locations of the object recognition or segmentation module, the pose estimation accuracy will also be affected. Then, since we use both synthetic and real images to train the network, there is a certain gap between these two types of images, which causes some learning difficulties during network training. Finally, there is still a gap in accuracy between RGB-based methods and RGB-D-based methods. In future work, we will focus on addressing these issues by using other object detection networks, improving the gap between synthetic and real images, and enhancing pose prediction accuracy.

\section{Conclusion}
In this paper, we propose a novel RGB-based category-level object pose estimation method. We first construct a transformer-based geometric feature prediction architecture and feature fusion method to fully utilize RGB information. Then, we propose a geometric guidance method that can effectively guide the network to accurately predict the object's geometric features. Finally, we conduct performance tests on the CAMERA25 and REAL275 datasets. Experimental results show that the proposed method can effectively complete category-level object pose estimation with only RGB images and achieve higher prediction accuracy.

%single-stage category-level multi-object pose tracking method based on convolution and vision transformer. We use a novel temporal transformer module to grasp the temporal information, and use a novel point matching method to group the center point and keypoints. The experiment results show that the CatTrack can track and handle multi-object tracking with a better performance, and get SOTA performance than other methods on Objectron dataset.

%There are also some weaknesses of CatTrack. Since the 3D model of the objecet is unknown, the object pose heavily rely on the prediction of the keypoint. When some of the keypoints are invisible, there will be some error between the prediction and ground truth. 

\bibliographystyle{IEEEtran}

% argument is your BibTeX string definitions and bibliography database(s)
\bibliography{IEEEabrv,paper_ref}

\end{document}